\documentclass{article}

\usepackage{arxiv}

\usepackage[utf8]{inputenc} 
\usepackage[T1]{fontenc}    
\usepackage{hyperref}       
\usepackage{url}            
\usepackage{booktabs}       
\usepackage{amsfonts}       
\usepackage{nicefrac}       
\usepackage{microtype}      
\usepackage{lipsum}
\usepackage{graphicx}
\usepackage{amsmath}
\usepackage{amssymb}
\usepackage{booktabs}
\usepackage{multirow}
\usepackage{subcaption}
\usepackage{pifont}
\newcommand{\cmark}{\ding{51}} 
\newcommand{\xmark}{\ding{55}} 

\title{Automated Landmark Detection \\for assessing hip conditions:\\A Cross-Modality Validation of MRI versus X-ray}
\author{
Roberto Di Via \\
  MaLGa Center, DIBRIS \\
  University of Genoa \\
  Genoa, Italy \\
  \And
Vito Paolo Pastore \\
  MaLGa Center, DIBRIS \\
  University of Genoa \\
  Genoa, Italy \\
  \And
Francesca Odone \\
  MaLGa Center, DIBRIS \\
  University of Genoa \\
  Genoa, Italy \\
  \And
Si\^{o}n Glyn-Jones \\
  Nuffield Department of Orthopaedics, Rheumatology \\
  and Musculoskeletal Sciences \\
  University of Oxford \\
  Oxford, UK \\
  \And
Irina Voiculescu\thanks{Correspondence to \texttt{irina@cs.ox.ac.uk}} \\
  Department of Computer Science \\
  University of Oxford \\
  Oxford, UK \\
}

\begin{document}
\maketitle
\begin{abstract}
Many clinical screening decisions are based on angle measurements. In particular, FemoroAcetabular Impingement (FAI) screening relies on angles traditionally measured on X-rays. However, assessing the height and span of the impingement area requires also a 3D view through an MRI scan.
The two modalities inform the surgeon on different aspects of the condition. 
In this work, we conduct a matched-cohort validation study (89 patients, paired MRI/X-ray) using standard heatmap regression architectures to assess cross-modality clinical equivalence. Seen that landmark detection has been proven effective on X-rays, we show that MRI also achieves equivalent localisation and diagnostic accuracy for cam-type impingement. Our method demonstrates clinical feasibility for FAI assessment in coronal views of 3D MRI volumes, opening the possibility for volumetric analysis through placing further landmarks.
These results support integrating automated FAI assessment into routine MRI workflows. Code is released at \href{https://github.com/Malga-Vision/Landmarks-Hip-Conditions}{https://github.com/Malga-Vision/Landmarks-Hip-Conditions}. 
\end{abstract}


%

\section{Introduction}
\label{sec:intro}

\begin{figure}[!ht]
    \centering
    \includegraphics[width=\linewidth]{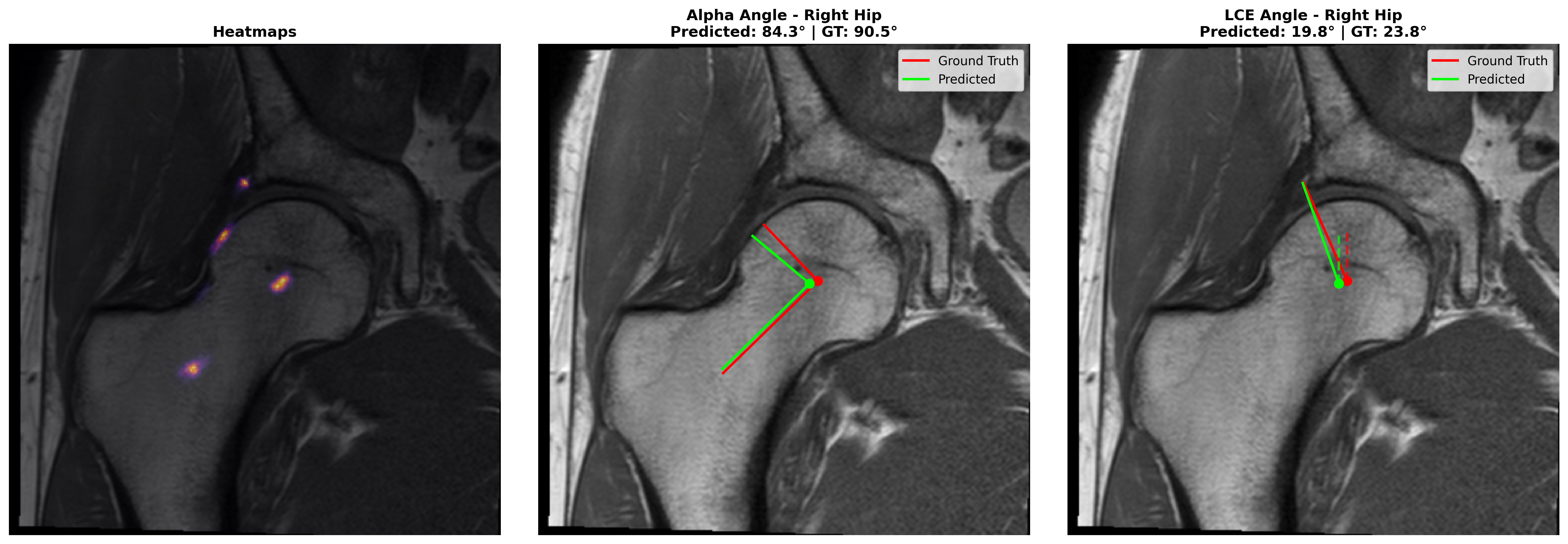}
    \includegraphics[width=\linewidth]{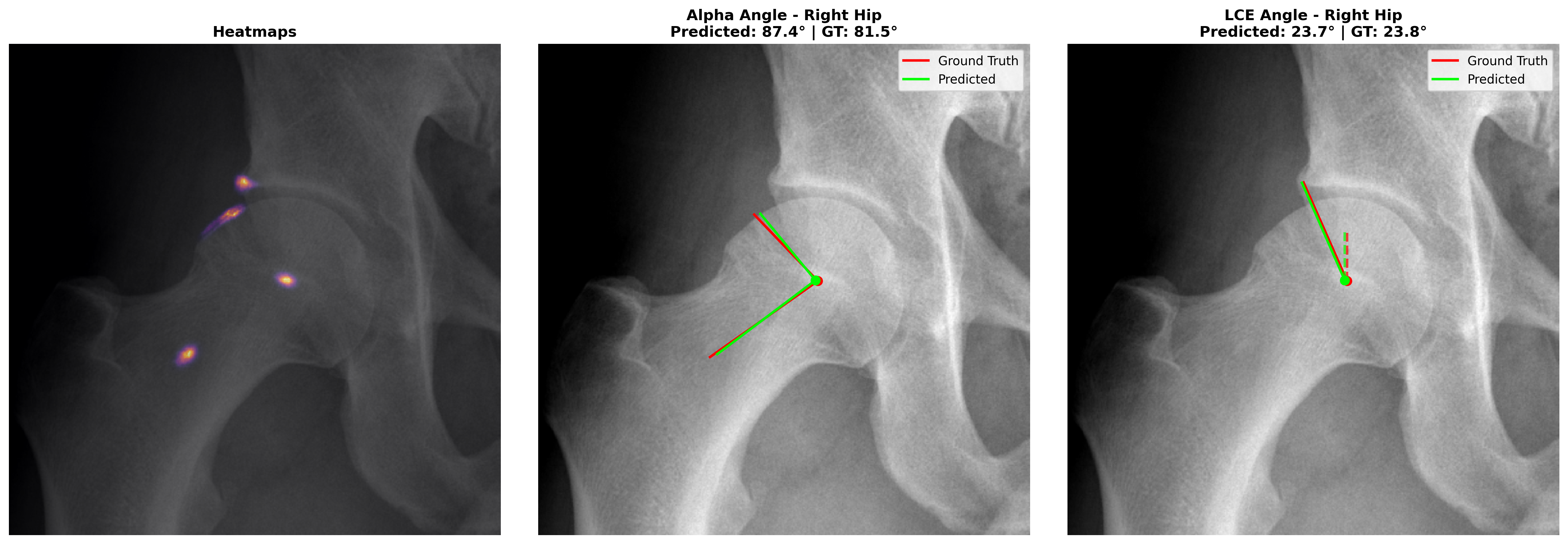}
    \caption{FAI assessment from paired T1 MRI (top) and X-ray (bottom). The $\alpha$-angle and LCE angle quantify femoral head asphericity and acetabular coverage, respectively. Our study confirms that automated landmark detection achieves equivalent accuracy across modalities, confirming the feasibility of MRI landmark detection.}
    \label{fig:teaser}
\end{figure}

Femoroacetabular impingement (FAI) is a pathomechanical hip disorder affecting 20–25\% of the population, characterised by abnormal contact between the femoral head–neck junction and acetabular rim during motion~\cite{ganz2003fai}. Early identification is critical, as untreated FAI accelerates degenerative cartilage loss and predisposes to premature osteoarthritis~\cite{agricola2013cam}.
FAI is the main cause of hip replacement. However, it is hard to diagnose at its early stages. Imaging is essential for early detection and treatment. Clinical assessment relies on two geometric parameters (Fig.~\ref{fig:teaser}): the $\alpha$-angle, quantifying femoral head asphericity (cam morphology), and the lateral centre-edge (LCE) angle, measuring acetabular coverage (pincer morphology). Pathological thresholds are typically $\alpha {>} 65^\circ$ and LCE${>} 40^\circ$~\cite{fraitzl2013femoral,kutty2012reliability}.

Current clinical practice relies on anteroposterior (AP) pelvic X-rays, which have fundamental limitations from tissue density being projected into a single view plane. The foot-to-hip alignment and pelvic tilt can affect the plane in which the relevant angles are being measured, leading to a variability of up to $15^{\circ}$~\cite{clohisy2008radiographic}, sufficient to misclassify borderline cases.
In this work, we investigate whether geometric FAI metrics can be reliably extracted from MRI alone. Whilst foot-to-hip misalignment can still be present in the coronal view, there are methods for correcting the misalignment, and measuring the angle in the correct plane. The first step towards this important 3D analysis is precisely showing that landmarks can be detected automatically in MRI slices.

Deep learning-based landmark detection have been used in diverse medical imaging tasks~\cite{clement2025confidence,robertoISBI,patel2025handful}, with recent work demonstrating the feasibility of an automated FAI assessment from X-rays~\cite{mccouat2021automatically}. However, no prior work has validated automated FAI assessment on MRI or directly compared cross-modality performance on matched cohorts. MRI offers intrinsic advantages over X-ray, because the alignment of the hip can be corrected post-scan, and the true 3D nature of the impingement area can be assessed independently, either across multiple slices separately or, ideally, through a single volumetric assessment.

\textbf{Contributions.} (1) We present the first cross-modality validation of automated hip landmark detection between X-ray and MRI. Standard heatmap regression models achieve equivalent performance on T1-weighted MRI (2.98 MRI vs.\ 3.02~mm X-ray mean error, 87.5\% diagnostic accuracy) on a matched patient cohort. This establishes coronal MRI as a successful modality for automated FAI assessment; (2) By extending measurements conventionally taken from X-rays to the corresponding slice in 3D MRI volumes, we enable seamless integration into clinical workflows; (3) We thereby establish a validated 2D foundation for future volumetric extensions towards full 3D geometric analysis and prediction.
\section{Method}
\label{sec:method}

\subsection{Problem Formulation}
\label{subsec:formulation}

We formulate FAI angle computation as a landmark detection problem. 
For each hip, we localise four annotated anatomical keypoints:  femoral head centre (FHC), neck-axis point (NA) along the femoral neck centreline,  lateral acetabular edge (LAE), and lateral cam point (LCP) where the femoral head deviates from sphericity (see Fig. \ref{fig:xray-mri-match}). 

The $\alpha$-angle is computed as the angle between the femoral neck axis (FHC$\rightarrow$NA) and the cam deformity vector (FHC$\rightarrow$LCP). Due to the LCP being hard to pinpoint -- even by clinicians -- $\alpha$  is notoriously difficult to calculate.

The easier LCE angle is calculated as the angle between the vertical axis and the acetabular coverage vector (FHC$\rightarrow$LAE), quantifying lateral acetabular overhang. Angles are visualised in Fig \ref{fig:teaser}.

\subsection{Heatmap Regression Architecture}
\label{subsec:architecture}

Following established practice in medical landmark detection~\cite{robertoWACV,payer2016heatmap}, we adopt a heatmap regression formulation. 
For each landmark $k \in \{1,\ldots,4\}$, we generate a ground-truth Gaussian heatmap $H_k{\in}\mathbb{R}^{h \times w}$, centred at the annotated location $(x_k, y_k)$ as 
$H_k(i,j) = \exp\!\left(-\frac{(i - y_k)^2 + (j - x_k)^2}{2\sigma^2}\right)$, 
with $\sigma{=}5$ empirically providing the best trade-off between localisation precision and training stability. 
Encoder–decoder networks are trained to predict heatmaps $\hat{H}_k$ from input images $I{\in}\mathbb{R}^{512 \times 512}$ by minimizing the negative log-likelihood (NLL) over the four landmarks~\cite{mccouat2022contour}.
At inference, landmark coordinates are extracted as the spatial argmax of each predicted heatmap. 
 For robustness, we employ test-time augmentation (TTA) by averaging heatmap predictions across different augmented views per test image, applying the  stochastic transformations described in Sec.~\ref{subsec:dataset}.

\section{Experiments}
\label{sec:experiments}

\subsection{Dataset and Training Setup}
\label{subsec:dataset}

We use a paired dataset of 89 patients who underwent AP pelvic X-ray and multiple hip MRI for FAI evaluation, collected at Oxford University as part of the FAIT trial ~\cite{palmer2014fait}. This pathological cohort includes subjects with clinical and imaging evidence of FAI but without significant osteoarthritis or hip dysplasia. Data were acquired using the same protocol, across multiple UK sites, with some patients contributing multiple time points. Since acquiring such a rich dataset, with multiple imaging modalities, longitudinally for the same patient is logistically and clinically challenging, the resulting paired dataset remains relatively small.

\begin{figure}[!t]
    \centering
    \includegraphics[width=\linewidth]{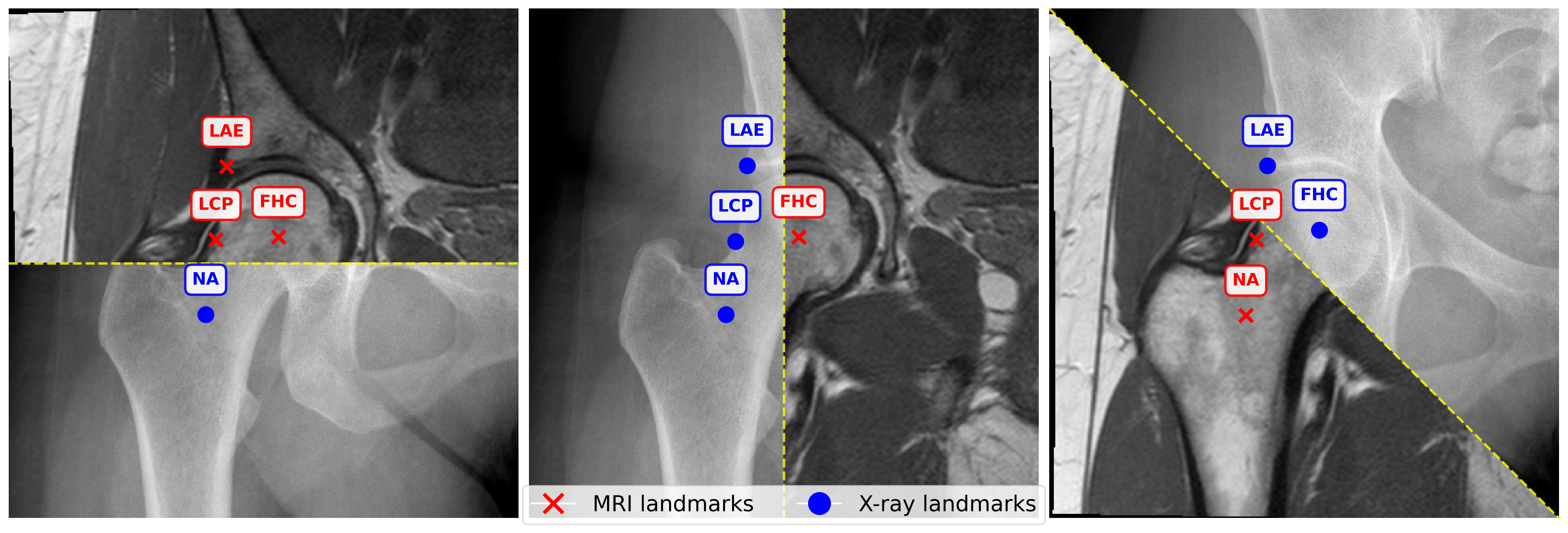}
    \caption{Example of paired anteroposterior pelvic X-ray and corresponding T1-weighted MRI (slice 10) from the same subject, showing annotated landmarks: femoral head centre (FHC), neck-axis point (NA), lateral acetabular edge (LAE), and lateral cam point (LCP), illustrating their spatial correspondence across modalities.}
    \label{fig:xray-mri-match}
\end{figure}

For MRI, we only used T1-weighted coronal acquisitions consisting of 20 slices with 3.3\,mm spacing. All images are resized/padded to $512{\times}512$ and min–max normalised to $[0,1]$.  All annotations are standardised to the middle slice (index 10), identified through pilot analysis of 20 subjects as the mid-acetabular coronal plane where all four landmarks (FHC, NA, LAE, LCP) are concurrently visible with maximal clarity.
Future work will investigate automatic slice selection to 
eliminate manual standardisation.
Example paired X-ray and MRI images with annotated landmarks are shown in Fig.~\ref{fig:xray-mri-match}. During training, stochastic augmentations include affine transforms (scale $0.95$–$1.05$, translation $\pm5\%$, rotation $\pm10^\circ$, shear $\pm5^\circ$) and intensity jitter (brightness/contrast $\pm15$–$20\%$, gamma $0.85$–$1.15$).  
The training:validation:test datasplit is $65{:}10{:}25$
balanced across $\alpha$-angle distributions given by Kolmogorov–Smirnov testing. This yields 105:17:40 images (57:8:24 patients).  

We train a UNet++ with a ResNet18 encoder (ImageNet-initialised) independently on the MRI and X-ray datasets. Optimisation uses AdamW (learning rate $1{\times}10^{-4}$, weight decay $1{\times}10^{-5}$). Training uses the NLL loss with an ExponentialLR scheduler ($\gamma=0.95$), a batch size of 4, and early stopping.

\subsection{Evaluation Metrics}
\label{subsec:metrics}

We evaluate performance across three hierarchical levels, and all metrics are computed independently for X-ray and MRI test sets. Experiment values are reported as mean ± standard deviation across three independent training runs with different random seeds.

\textbf{Landmark localisation.} We report (1) \textit{Mean Radial Error} (MRE): the average Euclidean distance between predicted and ground-truth landmarks (mm); (2) \textit{per-landmark MRE} for each keypoint ~\cite{robertoICCV}; and (3) \textit{Success Detection Rate} at radius $r$ (SDR@$r$): the percentage of landmarks localised within $r$\,mm.  

\textbf{Clinical angle assessment.} We measure (1) Mean Absolute Error (MAE) between predicted and ground-truth angles; (2) the Intraclass Correlation Coefficient (ICC(2,1)), a two-way random effects model quantifying absolute agreement by partitioning variance into between-subject variability, between-rater systematic bias, and residual error through ANOVA decomposition of the paired measurements (model vs.\ clinician). ICC ranges from 0 (no agreement) to 1 (perfect agreement), with values ${<}0.40$ indicating poor, $0.40$-$0.59$ fair, $0.60$-$0.74$ good, and ${>}0.75$ excellent clinical reliability~\cite{cicchetti1994guidelines}. (3) the median absolute difference as a robust summary; and (4) Bland–Altman analysis~\cite{bland1986statistical} to decompose disagreement into systematic bias, proportional bias, and random error. For each subject, we compute the difference and mean between predicted and ground-truth angles, reporting: (i) mean bias (systematic over- or under-estimation), (ii) 95\% limits of agreement (interval containing 95\% of differences), and (iii) proportional bias via linear regression testing whether error magnitude depends on angle value.

\textbf{Diagnostic performance.} For cam-type impingement detection ($\alpha{>}65^\circ$), we report accuracy, sensitivity, specificity, positive predictive value (PPV), and negative predictive value (NPV), together with the confusion matrix statistics.

\section{Results}
\label{sec:results}

\subsection{Landmark Localisation Performance}
\label{subsec:localisation}

Table~\ref{tab:localisation} summarises localisation accuracy. Per-landmark metrics report mean $\pm$ standard deviation over three runs; the `Overall' row shows per-run averages across landmarks, then aggregated across runs.
MRI achieves an overall MRE of 2.98$\pm$0.23~mm, closely matching that for X-rays (of 3.02$\pm$0.10~mm), demonstrating successful application of automated landmark detection to MRI. Individual landmark analysis reveals complementary strengths: X-rays excel at femoral head centre localisation (1.07~mm vs.\ 1.78~mm), likely due to higher bone-air contrast, while MRI shows superior neck-axis point detection (1.25~mm vs.\ 2.35~mm), potentially benefiting from improved soft-tissue delineation. Both modalities exhibit higher errors for the elusive lateral cam point (LCP: 5.50--5.72~mm mean), reflecting its clinical variability and frequent inter-observer disagreement; median errors (1.76--2.55~mm) are considerably lower, indicating that outliers in borderline FAI cases drive the means.
Success detection rates at clinical thresholds (2--4~mm) remain similar for MRI 58--83\% vs.\ X-ray 63--85\%.

\begin{table}[t] 
\centering 
\caption{Landmark localisation accuracy for X-ray and MRI modalities. Mean and median radial errors (RE) are reported for each landmark. SDR@r indicates the success detection rate within r mm. $\downarrow$ indicates lower is better; $\uparrow$ indicates higher is better.} 
\label{tab:localisation} 
\footnotesize 
\setlength{\tabcolsep}{5pt}
\resizebox{0.7\textwidth}{!}{
\begin{tabular}{@{}lcccc@{}} 
\toprule 
\textbf{Landmark} & \multicolumn{2}{c}{\textbf{Mean RE (mm)} $\downarrow$} & \multicolumn{2}{c}{\textbf{Median RE (mm)} $\downarrow$} \\ 
\cmidrule(lr){2-3} \cmidrule(lr){4-5} & X-ray & MRI & X-ray & MRI \\
\midrule 
FHC & \textbf{1.07$\pm$0.08} & 1.78$\pm$0.13 & 1.01$\pm$0.11 & \textbf{0.84$\pm$0.14} \\ 
NA & 2.35$\pm$1.02 & \textbf{1.25$\pm$0.30} & 1.61$\pm$0.17 & \textbf{1.13$\pm$0.36} \\ 
LAE & 3.18$\pm$0.64 & \textbf{3.17$\pm$0.30} & 2.81$\pm$0.74 & \textbf{2.78$\pm$0.09} \\ 
LCP & \textbf{5.50$\pm$1.01} & 5.72$\pm$0.76 & \textbf{1.76$\pm$0.66} & 2.55$\pm$0.64 \\ 
\midrule 
\textbf{Overall} & 3.02$\pm$0.10 & \textbf{2.98$\pm$0.23} & \textbf{1.49$\pm$0.16} & 1.62$\pm$0.29 \\ 
\midrule 
\multicolumn{5}{l}{\textbf{SDR@2/3/4mm (\%)} $\uparrow$ \textbf{X-ray:} 62.5/79.2/85.4 \quad \textbf{MRI:} 58.3/73.5/83.1} \\
\bottomrule 
\end{tabular} 
}
\end{table}

\subsection{Angle Agreement and Diagnostic Accuracy}
\label{subsec:angles}

Table~\ref{tab:angles_diagnostic} summarises the agreement in angle measurement and cam screening results. Both modalities agree strongly on the LCE-angle (X-ray ICC: 0.82, MRI ICC: 0.73, excellent clinical significance) with small median errors (2.22$^\circ$ and 1.28$^\circ$), indicating the geometric stability when localising the lateral acetabular edge.
In contrast, the $\alpha$-angle shows only moderate intra-modality agreement (ICC: 0.52 for X-ray, 0.41 for MRI; fair clinical significance) and larger MAEs (12.18$^\circ$, 13.64$^\circ$), comparable to the inter-observer variability among clinicians ($\approx$ 0.45--0.56)~\cite{ewertowski2022automated}, reflecting the inherent ambiguity of the cam deformity point. 
Median errors remain clinically acceptable (5.61$^\circ$, 5.45$^\circ$), suggesting that outliers inflate the mean disproportionately.
Although continuous $\alpha$-angle agreement differs modestly, both modalities achieve the same diagnostic accuracy for cam-type impingement (87.5\%) considering the best runs. They have perfect specificity (100\%) and equal sensitivity (54.6\%), indicating that the remaining variability lies within clinically acceptable limits.

\begin{table}[t]
\centering
\caption{Comparison of angle measurement agreement and diagnostic accuracy between X-ray and MRI. Results are reported as mean $\pm$ standard deviation across three different runs.}
\label{tab:angles_diagnostic}
\footnotesize
\setlength{\tabcolsep}{5pt}
\resizebox{0.6\textwidth}{!}{

\begin{tabular}{@{}lcc@{}}
\toprule
\textbf{Metric} & \textbf{X-ray} & \textbf{MRI} \\
\midrule
\multicolumn{3}{@{}l@{}}{\textit{LCE-angle agreement}} \\
MAE ($^\circ$) $\downarrow$   & 2.97$\pm$1.14 & \textbf{2.96$\pm$0.52} \\
Median ($^\circ$) $\downarrow$ & 2.22$\pm$0.42 & \textbf{1.28$\pm$0.20} \\
ICC (2,1) $\uparrow$          & \textbf{0.82$\pm$0.20} & 0.73$\pm$0.14 \\
\midrule
\multicolumn{3}{@{}l@{}}{\textit{$\alpha$-angle agreement}} \\
MAE ($^\circ$) $\downarrow$   & \textbf{12.18$\pm$1.63} & 13.64$\pm$1.77 \\
Median ($^\circ$) $\downarrow$ & 5.61$\pm$1.41 & \textbf{5.45$\pm$0.77} \\
ICC (2,1) $\uparrow$          & \textbf{0.52$\pm$0.06} & 0.41$\pm$0.15 \\
\midrule
\multicolumn{3}{@{}l@{}}{\textit{Cam screening $\alpha>65^\circ$}} \\
Accuracy (\%) $\uparrow$      & 87.50 & 87.50 \\
Sensitivity / Specificity (\%) $\uparrow$ & 54.55 / 100.00 & 54.55 / 100.00 \\
PPV / NPV (\%)                & 100.00 / 85.29 & 100.00 / 85.29 \\
TP / FP / TN / FN             & 6 / 0 / 29 / 5 & 6 / 0 / 29 / 5 \\
\bottomrule
\end{tabular}
}
\end{table}

\subsection{Bland--Altman Agreement Analysis}
\begin{figure}[!ht]
    \centering
    \includegraphics[width=0.9\linewidth]{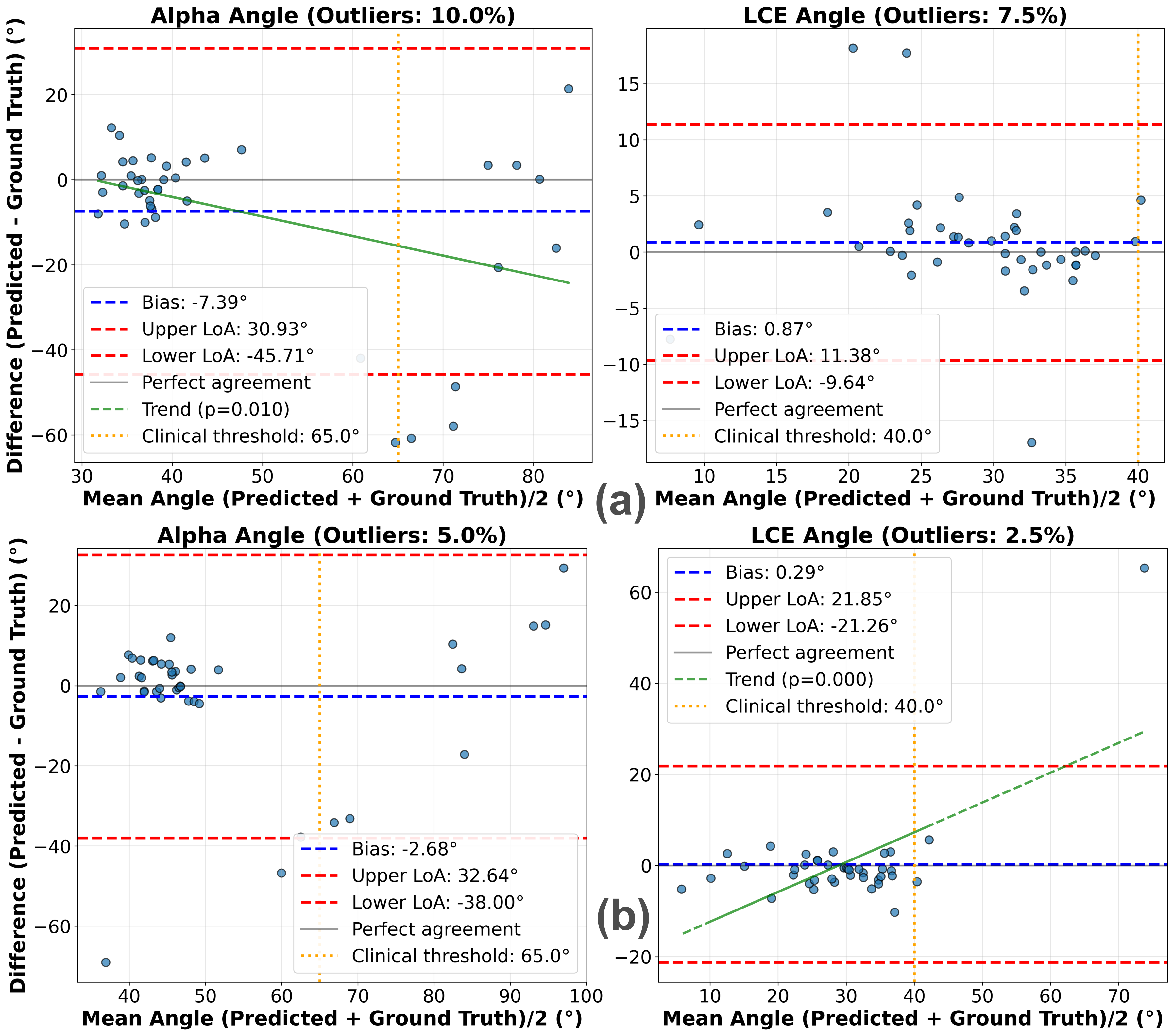}
    \caption{Bland--Altman plots comparing predicted and ground-truth $\alpha$- (left) and LCE-angles (right) for MRI (a) and X-ray (b).}
    \label{fig:bland-altman}
\end{figure}

To complement the ICC results and assess continuous measurement agreement, we perform Bland--Altman analysis comparing predicted and ground-truth angles (Fig.~\ref{fig:bland-altman}). 
Each plot shows the mean of the two measurements on the $x$-axis and their difference on the $y$-axis; dashed blue line indicates mean bias and red ones mark the 95\% limits of agreement (LoA). Points near zero bias and within the LoA indicate good agreement, while a regression trend denotes proportional bias.
For the LCE-angle, MRI shows minimal bias ($0.87^\circ$) and narrow LoA ($\pm10.5^\circ$) without proportional bias ($p=0.304$). In contrast, X-ray shows wider LoA ($\pm21.6^\circ$) and proportional bias ($p<0.001$), with errors increasing at higher acetabular coverage, suggesting more stable estimation from MRI. 
For the $\alpha$-angle, both modalities underestimate clinician measurements (X-ray: $-2.68^\circ$, MRI: $-7.39^\circ$) with broad LoA (X-ray: $\pm35.3^\circ$, MRI: $\pm38.3^\circ$). X-ray errors remain magnitude-independent ($p{=}0.518$), whereas MRI shows proportional bias ($p{=}0.010$), reflecting slice-selection sensitivity. Despite these biases, both modalities achieve reliable cam-type classification (specificity 100\%, sensitivity 54.6\%), indicating that residual variability mainly arises from landmark ambiguity rather than modality differences.

\subsection{Ablation Studies}
\label{subsec:ablation}

Table~\ref{tab:ablation} reports the effects of architecture choices and test-time augmentation (TTA) on the MRI results. 
To identify optimal network configurations, we compare established architectures (UNet++, UNet, DPT) with both convolutional (ResNet18, VGG16) and transformer-based (MiT-B1, MaxViT-Base) encoders of similar parameter counts (11--15M for lightweight, 119M for MaxViT-Base). All the encoders are ImageNet-initialised.
Our objective is not architectural novelty but rather systematic validation of which existing designs transfer best to MRI-based FAI assessment.
UNet++ with ResNet18 achieves optimal performance (MRE: 2.98$\pm$0.23~mm, mean cam accuracy: 84.75$\pm$2.77\%), outperforming standard UNet (3.98~mm, 69.83\%) and DPT variants (3.63--4.44~mm, 70--77.5\%). Lighter encoders (ResNet18) outperform heavier alternatives, suggesting moderate-capacity networks with multi-scale skip connections are well-suited for this task. TTA consistently improves all architectures, providing 1.4--3.6 percentage point gains in cam accuracy while reducing variance. X-ray results show similar trends.

\begin{table}[t]
\centering
\footnotesize
\renewcommand{\arraystretch}{1.05}
\caption{Ablation study of architecture–encoder combinations and test-time augmentation (TTA) on MRI performance, reporting mean radial error (MRE, mm), success detection rate at 2 mm (SDR@2 mm), and cam-type diagnostic accuracy (Cam Acc.).}
\label{tab:ablation}
\resizebox{0.8\textwidth}{!}{
\begin{tabular}{@{}l c c c c@{}}
\toprule
\textbf{Model} & \textbf{TTA} & \textbf{MRE (mm)} $\downarrow$ & \textbf{SDR@2mm (\%)} $\uparrow$ & \textbf{Cam Acc. (\%)} $\uparrow$ \\
\midrule
UNet++ (ResNet18)       & \xmark & $3.02 \pm 0.28$ & $56.46 \pm 5.87$ & $82.33 \pm 3.79$ \\
UNet++ (ResNet18)       & \cmark & $\mathbf{2.98 \pm 0.23}$ & $\mathbf{58.44 \pm 4.86}$ & $\mathbf{84.75 \pm 2.77}$ \\
UNet++ (VGG16)          & \xmark & $3.33 \pm 0.03$ & $56.25 \pm 0.89$ & $80.00 \pm 7.07$ \\
UNet++ (VGG16)          & \cmark & $3.29 \pm 0.36$ & $57.09 \pm 1.93$ & $80.17 \pm 4.41$ \\
\midrule
UNet (MiT-B1)           & \xmark & $4.11 \pm 0.84$ & $54.79 \pm 1.44$ & $66.67 \pm 12.58$ \\
UNet (MiT-B1)           & \cmark & $\mathbf{3.98 \pm 0.54}$ & $\mathbf{57.33 \pm 0.91}$ & $\mathbf{69.83 \pm 6.85}$ \\
\midrule
DPT (ResNet18)          & \xmark & $4.49 \pm 0.54$ & $48.34 \pm 3.79$ & $67.50 \pm 2.50$ \\
DPT (ResNet18)          & \cmark & $4.44 \pm 0.19$ & $50.21 \pm 3.02$ & $70.17 \pm 6.56$ \\
DPT (MaxViT-Base)       & \xmark & $3.91 \pm 0.87$ & $55.50 \pm 0.71$ & $73.75 \pm 8.84$ \\
DPT (MaxViT-Base)       & \cmark & $\mathbf{3.63 \pm 0.99}$ & $\mathbf{55.94 \pm 1.33}$ & $\mathbf{77.50 \pm 10.61}$ \\
\bottomrule
\end{tabular}
}
\end{table}

\section{Conclusion \& Future Work}
\label{sec:conclusion}

Our study focuses on patients with FAI, whose diagnosis and management, including surgical planning, depend on an accurate two-dimensional and volumetric assessment of the impingement. By demonstrating that automated landmark detection on MRI achieves accuracy equivalent to that on X-rays, we establish that geometric measures of FAI can be extracted reliably from routine MRI scans. This represents an important first step towards supporting MRI as a suitable modality for automated FAI assessment. 

Although AP X-rays remain the first-line tool for clinical screening due to their speed, low cost, and accessibility, they are not ideal for surgical planning. Our findings highlight the potential of MRI for quantitative geometric assessment in musculoskeletal analysis. Beyond the demonstrated equivalence, the volumetric nature of MRI offers opportunities for refined development of screening angles, particularly in patients with asymmetries or pelvic rotations, where measurements must be made relative to local axes. 

Current limitations relate to analysing a single MRI slice, which underutilises volumetric information and disregards inter-slice coherence. The moderate $\alpha$-angle reliability is due to the cam-point ambiguity and the modest size of the cohort. 

Future work will extend the approach towards volumetric metrics averaged across slices or computed directly from 3D MRI volumes, exploring adjacent-slice training for data augmentation and incorporating uncertainty quantification to identify unreliable predictions.

This research was conducted retrospectively using human subject data from the FAIT study~\cite{palmer2014fait}, which received ethical approval. The authors have no conflicts of interest to declare.

{
\bibliographystyle{unsrt}
\bibliography{refs}
}

\end{document}